\pdfoutput=1
\documentclass[11pt]{article}

\usepackage[final]{acl}

\usepackage{times}
\usepackage{latexsym}

\usepackage[T1]{fontenc}
\usepackage[utf8]{inputenc}

\usepackage{microtype}

\usepackage{inconsolata}

\usepackage{graphicx}
\usepackage{subcaption}
\title{Chain of Methodologies: Scaling Test Time Computation without Training}

\author{Cong Liu, Jie Wu$^\dagger$, Weigang Wu, Xu Chen, Liang Lin, Wei-Shi Zheng \\
  Sun Yat-sen University, $^\dagger$Temple University \\
  \texttt{liucong3@mail.sysu.edu.cn, jiewu@temple.edu, wuweig@mail.sysu.edu.cn,} \\ \texttt{chenxu35@mail.sysu.edu.cn, linliang@ieee.org, wszheng@ieee.org} \\}

\begin{document}
\maketitle

\begin{abstract}

Large Language Models (LLMs) often struggle with complex reasoning tasks due to insufficient in-depth insights in their training data, which are typically absent in publicly available documents. This paper introduces the Chain of Methodologies (CoM), an innovative and intuitive prompting framework that enhances structured thinking by integrating human methodological insights, enabling LLMs to tackle complex tasks with extended reasoning. CoM leverages the metacognitive abilities of advanced LLMs, activating systematic reasoning throught user-defined methodologies without explicit fine-tuning. Experiments show that CoM surpasses competitive baselines, demonstrating the potential of training-free prompting methods as robust solutions for complex reasoning tasks and bridging the gap toward human-level reasoning through human-like methodological insights.

\end{abstract}

\section{Introduction}

\begin{figure}[t]
  \includegraphics[trim=115 90 370 90,clip,width=0.48\textwidth]{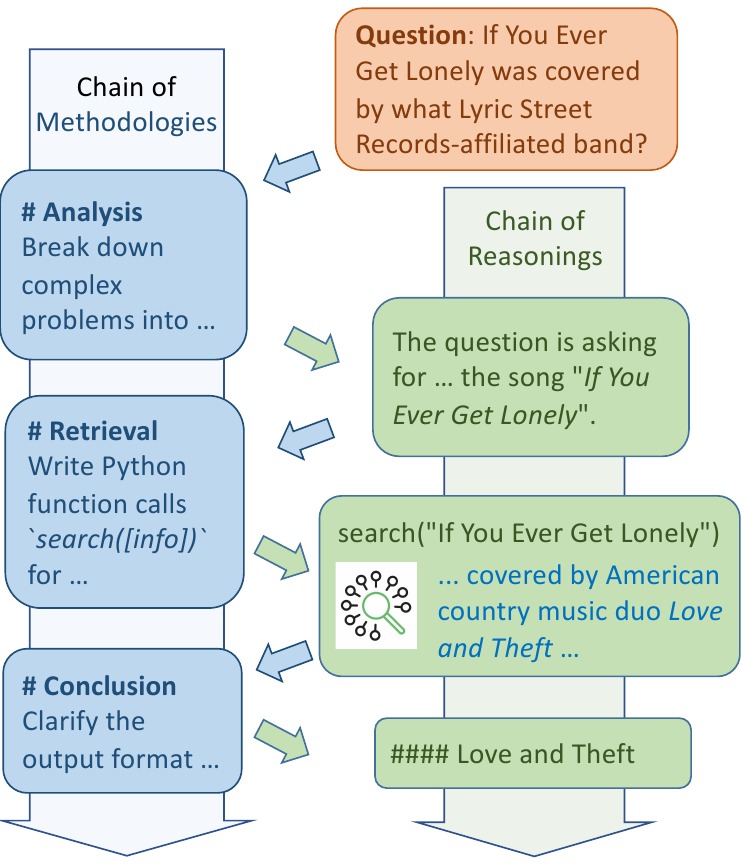}
  \caption{An Example of our Chain of Methodologies reasoning process, where the generation of methodologies and reasoning interleaves. A methodology (in blue) is selected based on the historical reasoning status, while the next reasoning step (in green) is guided by the previously selected methodology.}
  \label{fig:overview}
\end{figure}

Recently, OpenAI's o1 \cite{o1} showcases the possibility of using a long chain of thoughts to improve the reasoning ability of Large Language Models (LLMs). During these long thoughts, OpenAI's o1 displays high-level cognitive abilities, such as problem decomposition, error recognition, and correction, which constantly steer the thoughts in the right direction. OpenAI confers o1 with such abilities through reinforcement learning.

This paper explores whether LLMs can achieve similar self-guiding abilities for long, structured reasoning across domains using only prompts, without instruction fine-tuning. This is a challenging problem: while fine-tuning with large datasets can broadly improve instruction-following, conventional prompts are typically limited to specific tasks with few-shot examples due to constraints like context length and information extraction accuracy. As a result, pure prompting methods are rarely used for cross-domain tasks, despite their advantages—low cost, rapid deployment, high sample efficiency, and avoidance of catastrophic forgetting or data bias.

Our work is inspired by prior research on \emph{metacognitive knowledge} in LLMs, which refers to the ability to reason about one's own reasoning processes. Pedagogical studies show that enhancing metacognitive knowledge improves reasoning in humans, and similar benefits have been observed in LLMs through prompts encouraging introspection and self-reflection \cite{metacognitive-prompt}. Besides, Microsoft's Phi-3 \cite{phi3} uses system prompts like "do not hallucinate" to reduce hallucination, while \cite{metacognitive-math} shows improved mathematical reasoning when LLMs identify required skills to retrieve relevant examples. These findings provide both intuitive and empirical support for our approach.

We propose the Chain of Methodologies (CoM), an intuitive task-agnostic prompting technique designed to enable cross-domain self-guided reasoning without instruction fine-tuning. CoM uses methodology as catalysts to stimulate LLMs to generate the next reasoning step based on the reasoning history. While LLMs often struggle with complex reasoning tasks due to insufficient in-depth insights between problems and their respective solutions in the training data, CoM bridges this gap and enables smooth transitions from a problem to its solution by inserting a methodological analysis before each solution step. This leverages the metacognitive knowledge of LLMs to select or generate methodologies that justify or explain the next steps.

\begin{figure*}[t]
  \centering
  \includegraphics[trim=45 368 398 80,clip,width=1\textwidth]{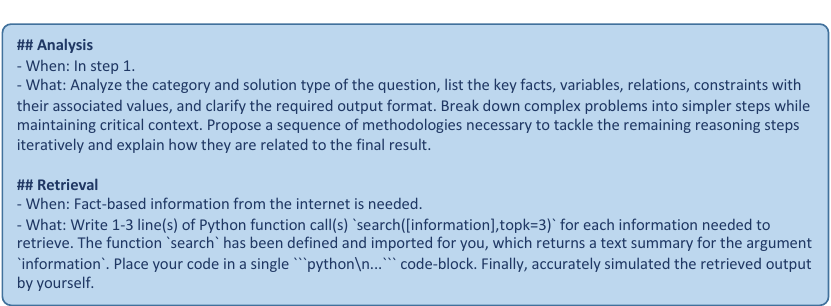}
  \caption{Two example methodology definitions in our \emph{when-what} format.}
  \label{fig:methodologies2}
\end{figure*}

CoM features two key components: (1) a list of methodologies formatted in our ``when-what'' format, which facilitates selection based on the reasoning history and connects it to the next reasoning step, and (2) a methodology-reasoning loop that iteratively selects the next methodology to guide reasoning along an extended and well-structured reasoning path. An example of CoM's interleaving methodology-selecting and reasoning path is illustrated in Figure~\ref{fig:overview}. Two examples of user-defined methodologies are listed in Figure~\ref{fig:methodologies2}.

Our contributions include the simple CoM framework and extensive experiments. CoM produces structured, explainable, and faithful reasoning paths. It is also highly extensible in that users can enhance the framework by modifying the list of methodologies in plain text. We evaluated our task-agnostic CoM framework on two types of representative and challenging tasks: mathematical reasoning and retrieval-augmented generation. Experiments show that CoM outperforms competitive baselines on these tasks across diverse LLMs.

\begin{figure}[t]
  \centering
  \includegraphics[trim=505 168 77 118,clip,width=0.35\textwidth]{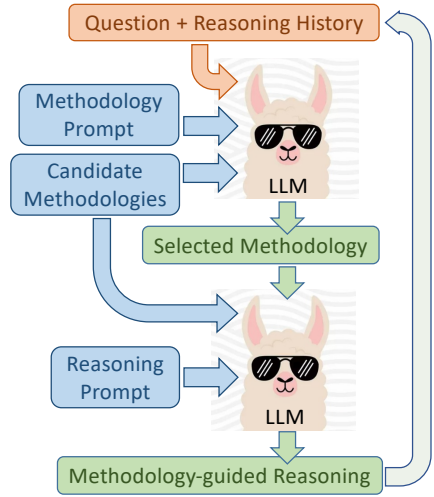}
  \caption{The components in our CoM framework and their interactions in the methodology-reasoning iterations.}
  \label{fig:overview2}
\end{figure}

\section{Chain of Methodologies}

\subsection{Overview}

We aim to use prompts to stimulate high-level cognitive (metacognitive) knowledge in existing LLMs, enabling them to possess the same cross-domain self-guiding ability as OpenAI's o1 thereby successfully carrying out extended and structured reasoning sequences across various domains. These prompts should be task-agnostic and effective in guiding thought processes. We find prompts related to methodology are ideal candidates for this purpose. Methodology is a critical component of any discipline or field that requires a structured approach to understanding, problem-solving, or conducting research. It provides a framework that ensures tasks are executed consistently and effectively. 

Our Chain of Methodologies (CoM) framework consists of a list of user-defined methodologies and a methodology-reasoning iteration. Each methodology provides a guideline for the next reasoning step based on the reasoning history. The reasoning process of CoM alternates between a methodology selection step and a methodology-guided reasoning step, as illustrated in Figure~\ref{fig:overview}.

\textbf{List of Methodologies}: Unlike AlphaGo, which operates within a defined set of rules and a closed action space, the general problem-solving ability of human in an open action space is more complex and challenging to optimize. In fact, the accumulation and evolution of human methodologies have relied on fundamental processes such as trial and error, reflection, and self-correction based on problem-solving experiences across different eras and civilizations. To navigate this complexity, we integrate human knowledge and experience related to task completion through established methodologies. Let $M = \{m^{(1)}, m^{(2)}, \cdots, m^{(n)}\}$ denote the list of $n$ user-defined methodologies.

\textbf{Reasoning iterations}: CoM conduct a maximum of $K$ steps for each question $Q$. In step $k$, where $1 \le k \le K$, we first prompt an $LLM_m$ with prompt template $P_p$ to select a methodology $m_k \in M$ based on the reasoning history $h_k$:
\begin{equation}
m_k = LLM_p(M, Q, h_k, P_p)
\end{equation}
, and then prompt an $LLM_r$ with prompt template $P_r$ to generate the next reasoning sequence $r_k$  based on methodology $m_k$ and history $h_k$:
\begin{equation}
r_k = LLM_r(M, Q, h_k, m_k, P_r)
\end{equation}
, where the reasoning history contains all previous reasoning sequences $h_k = [r_1, r_2, \cdots, r_{k-1}]$. In this paper, we simply use the same instruction fine-tuned LLM for both $LLM_m$ and $LLM_r$, which is frozen during the application of our framework.

\subsection{Methodology Definition}

Our emphasis is on a framework that utilizes a user-defined list of methodologies rather than studying the philosophy of finding a universally applicable set of methodologies, whose existence is a debated topic between universalism and contextualism. From a pragmatic standpoint, we focus on representing each methodology in a way that facilitates methodology selection and methodology-based reasoning.

To clarify the distinction between method and methodology, a method refers a specific technique or systematic procedure for accomplishing a task, whereas a methodology encompasses the principles and rationale guiding the selection and appliction of methods. Each methodology in our user-defined list specifies two key fields: \emph{when} and \emph{what}. The \emph{when} field defines the applicable stage of the methodology in the reasoning lifecycle, along with the context and factors influencing the choice of the methodology. The \emph{what} field outlines the systematic approach, action selection criteria, and expected outcomes of the methodology.

Specifically, a methodology is defined in Markdown format with three fields: (1) its name, (2) \emph{when}: the situation and timing for its application, and (3) \emph{what}: its specification and details, including principles, tools, techniques, and procedures. Figure~\ref{fig:methodologies2} provides examples of two methodology definitions.

Next, we discuss different types of methodologies. We categorize methodology definitions into three broad types: \emph{analysis}, \emph{coding}, and \emph{reflection}. \emph{Analysis} methodologies guide the LLM in organizing information, such as extracting facts, variables, relations, constraints, and objectives from the question; breaking down the initial question into manageable sub-problems; planning the sequence of actions; and summarizing, rearranging, and distilling the information obtained so far. \emph{Coding} methodologies prompt the LLM to generate formal languages for execution by solvers to obtain accurate results, or to use external tools (e.g., search engines) by calling predefined functions attached to the solvers. \emph{Reflection} methodologies encourage the LLM to identify errors and provide constructive feedback through self-reflection or self-verification, enabling adjustments to the approach and proposing alternative strategies for subsequent steps. Figure~\ref{fig:methodologies} in Appendix~\ref{appendix:prompts} lists the task-agnostic methodology definitions we used in our experiments.

In summary, the use of methodologies serves a multifaceted purpose: (1) providing human-input methodologies to stimulate the metacognitive ability of LLMs, compensating for the lack of in-depth insights in their training data for complex reasoning task; (2) establishing a natural connection through explanation or justification between the current reasoning situation and its solution in the next step; and (3) offering an educated guess for the next step, avoiding the formidable complexity of stochastic search methods like MCTS \cite{rStar} and RL \cite{o1, scaling-testtime, star}, which operate over a general reasoning space that is much larger than those in games like AlphaGo.

Finally, our framework is designed for easy extensibility: users can update the list of methodology definitions in plain text to make it more comprehensive for general thinking or tailor it to a specific set of skills that accurately target a particular task.

\subsection{Methodology-Reasoning Iterations}

As illustrated in Figure~\ref{fig:overview2}, CoM alternates between prompting the LLM to generate the next methodology and the next methodology-based reasoning sequence for a maximum of $K$ iterations. 

The first prompt instructs the LLM to select a methodology for the next reasoning steps. This prompt concatenates the list of user-defined methodology definitions, the question, the history of previous methodology-based reasoning sequences, and an instruction that provides additional information about the reasoning stage and the output format. It enables the LLM to choose the most suitable methodology for the task.

The second prompt includes all the information from the first prompt, along with the methodology selected using the first prompt. It directs the LLM to adhere the guidance outlined in the chosen methodology while reasoning. Additionally, the second prompt requires the output to include the following elements: (1) an acknowledgment of the selected methodology by restating its name, (2) a chain-of-thought reasoning process or a program that implements the methodology, and (3) a summarized result of the reasoning or a guessed output of the program.

Following the second prompt, a solver is invoked to post-process the LLM's output. This step facilitates the LLM's programming ability \cite{PoT}. Currently, we have only implemented a Python interpreter, which is triggered when Python code blocks are detected in the output. This interpreter executes the code in a secure environment with several common packages pre-imported. After execution, the predicted output of the program in the LLM's response is replaced with the actual stdout output from the code's execution. This approach ensures accurate reasoning on tasks that require computation, such as mathematical tasks, effectively implementing the human methodology: ``You should use a calculator for tasks that involve complex calculations.'' Furthermore, it enables various types of tool-using via Python APIs during the reasoning process, including web searches, knowledge base retrieval, and even invocation of other LLMs or manipulation of the LLM's own reasoning process \cite{prob-tree-of-thought}.

Our Python interpreter executes code in a sandboxed environment, which operates as a new process with a safe global scope. In this environment, the code can only access a limited set of built-in functions and import from a predefined list of packages. We enforce a timeout of 1 minute for each process, as we empirically determined that larger timeouts do not significantly improve performance on our experimental tasks. Users can extend the tool-using capabilities of the CoM framework by adding corresponding methodology definitions and implementing relevant functions in the Python interpreter. For instance, to enable Google search, one could add a methodology definition specifying the existence of a function named ``search'' and the meaning of its arguments, followed by implementing and adding this function to the global scope of the Python interpreter.

Our prompts for methodology selection and methodology-based reasoning are provided in Figure~\ref{fig:select} in the Appendix~\ref{appendix:prompts}.

\section{Related Work}

\textbf{Prompting} A significant body of work has explored various prompt designs to enhance the reasoning capabilities of LLMs. Notable approaches include Chain-of-Thought \cite{wei2022chain}, Least-to-Most \cite{zhou2023leasttomost}, Self-Consistency \cite{wang2023selfconsistency}, and Tree-of-Thoughts \cite{prob-tree-of-thought}. Methods to enhance problem-specific performance, include question rephrasing, dividing subtasks, verification, symbolic grounding \cite{faithful-chain-of-thought, symbolic-chain-of-thought, plan-and-solve, star, CoK}, factuality and faithfulness verification for reasoning chains \cite{CoK}, as well as explicit separation of knowledge retrieval and reasoning steps to organize decision-making \cite{disentangle}.

\noindent \textbf{Iterative Prompting} Prior research has also investigated iterative prompting methods to structure reasoning processes. Examples include Self-Refine \cite{self_refine}, IRCoT \cite{IRCoT}, iCAP \cite{iCAP}, MetaGPT \cite{MetaGPT}, and Chain of Ideas \cite{CoI}. These approaches typically rely on predefined, hardcoded actions to guide reasoning. In contrast, our work introduces a task-agnostic framework that leverages the metacognitive abilities of LLMs to dynamically select methodologies based on reasoning history. Furthermore, while prior work focuses on generating the next reasoning step, our approach adopts a justification-before-action style, where the model introspectively justifies why a specific methodology is needed before executing it. This mirrors human metacognitive processes and distinguishes our work from implicit context-aware token generation.

\noindent \textbf{Metacognition-based} Several contemporary works are closely related to our approach. Buffer of Thoughts \cite{buffer-of-thoughts} derives high-level guidelines from previously completed tasks and stores them in a buffer for future reuse, enabling learning from experience and improving efficiency by distilling level-2 slow thinking into level-1 fast thinking. However, unlike our work, its high-level guidelines contain problem-specific reasoning chains or code templates tailored to particular tasks, such as complex multi-query tasks. Skill-based CoT \cite{metacognitive-math} explores the metacognitive capabilities of LLMs in mathematical problem-solving by labeling questions with corresponding skills, clustering them to reduce redundancy, and retrieving skill-relevant examples for in-context learning during inference. Induction-augmented generation \cite{induction-augmented-generation} identifies key concepts in questions and uses inductive prompting templates to extract their close concepts and common attributes, facilitating more accurate reasoning processes.

\noindent \textbf{Search-based} rStar \cite{rStar} introduces a self-play mutual reasoning approach that significantly improves the reasoning capabilities of small language models without fine-tuning. This method employs a costly Monte Carlo Tree Search (MCTS) with a set of five reasoning-inducing prompts.

\noindent \textbf{Training-based} Finally, training-based methods have been developed to enable LLMs to handle long chains of thought. For example, STaR \cite{star} demonstrates that iterative training on reasoning histories leading to correct answers enables models to solve increasingly complex problems. Similarly, \cite{scaling-testtime} fine-tunes small models to perform more reasoning steps using reinforcement learning with beam search, lookahead search, and best-of-N verifiers. ReST-MCTS \cite{ReST-MCTS} integrates process reward guidance with tree search MCTS to collect higher-quality reasoning traces, while AFlow \cite{aflow} iteratively refines task-specific workflows. These methods highlight the potential of training-based approaches but often require significant computational resources.

\begin{figure*}[ht]
\begin{subfigure}{.32\textwidth}
  \centering
  \includegraphics[width=\linewidth]{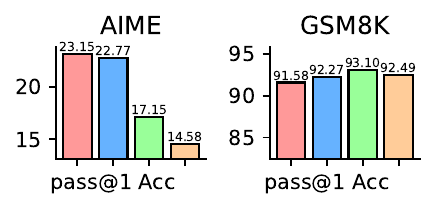}  
  \label{fig:qwen1}
\end{subfigure}
\begin{subfigure}{.68\textwidth}
  \centering
  \includegraphics[width=\linewidth]{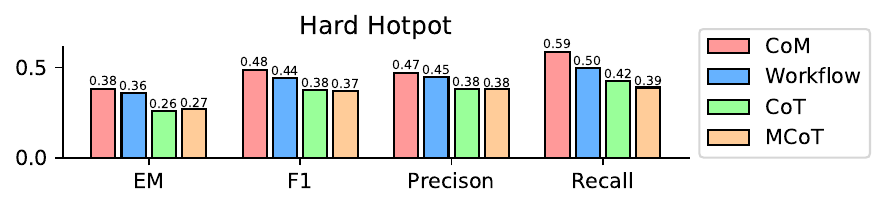}  
  \label{fig:qwen1}
\end{subfigure}
\vspace{-35pt}
\caption{Results of Qwen2-72B-Instruct on AIME, GSM8K, and Hard HotpotQA.}
\label{fig:qwen-large}
\end{figure*}

\begin{figure*}[ht]
\begin{subfigure}{.32\textwidth}
  \centering
  % include first image
  \includegraphics[width=\linewidth]{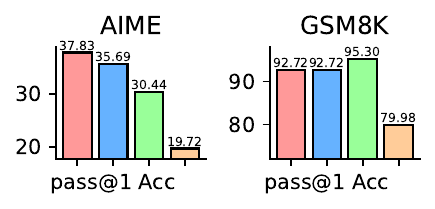}  
  \label{fig:qwen1}
\end{subfigure}
\begin{subfigure}{.68\textwidth}
  \centering
  % include second image
  \includegraphics[width=\linewidth]{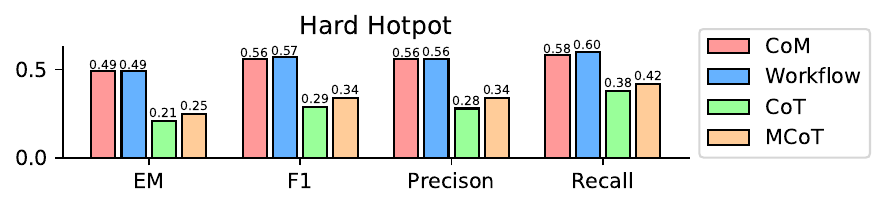}  
  \label{fig:qwen1}
\end{subfigure}
\vspace{-35pt}
\caption{Results of DeepSeek-V3 on AIME, GSM8K, and Hard HotpotQA.}
\label{fig:deepseek}
\end{figure*}

\begin{figure*}[ht]
\begin{subfigure}{.44\textwidth}
  \centering
  % include first image
  \includegraphics[width=\linewidth]{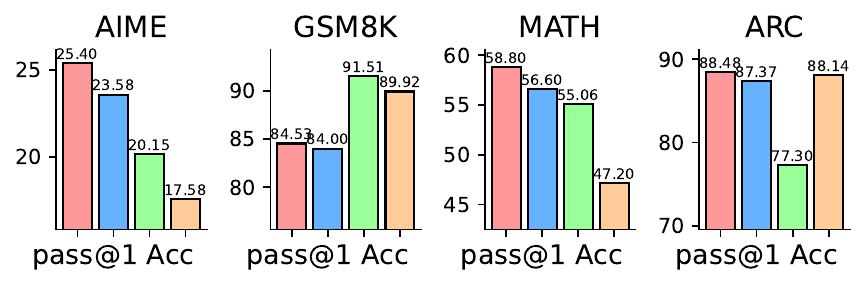}  
  \label{fig:qwen3}
\end{subfigure}
\begin{subfigure}{.56\textwidth}
  \centering
  % include second image
  \includegraphics[width=\linewidth]{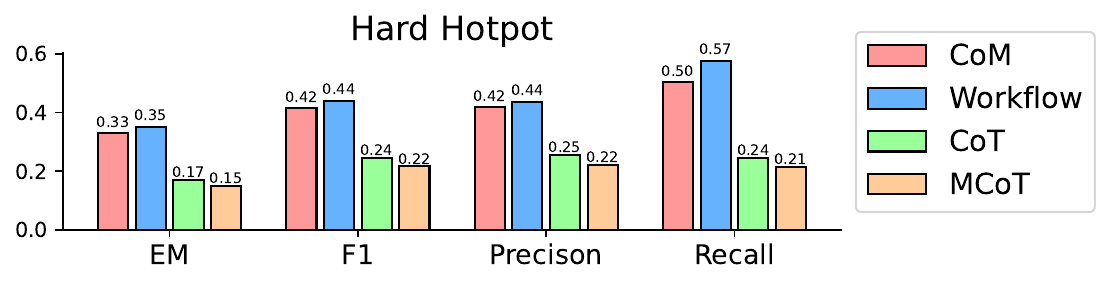}  
  \label{fig:qwen4}
\end{subfigure}
\vspace{-35pt} 
\caption{Results of Qwen2.5-7B-Instruct, a small LLM, on AIME, GSM8K, MATH, ARC, and Hard HotpotQA.}
\label{fig:qwen-small}
\end{figure*}

\section{Experiments}

\subsection{Experiment Setup}

\begin{table}[t]
    \small
    \centering
    \caption{LLMs used in our experiments. Results on the last three LLMs are reported in Appendix~\ref{appendix:fail}.}
    \begin{tabular}{l l} 
       \hline
       \textbf{LLM} & \textbf{Size} \\ 
       \hline
       DeepSeek-V3 \cite{deepseekv3} & 671B \\ 
       Qwen2-72B-Instruct \cite{qwen2} & 72B \\
       Qwen2.5-7B-Instruct \cite{qwen2} & 7B \\
       Macro-o1 \cite{marco-o1} & 7B \\
       Yi-1.5-9B-Chat \cite{yi} & 9B \\
       InternLM2.5-7B-chat \cite{cai2024internlm2} & 7B \\
       GLM-4-9b-chat \cite{glm2024chatglm} & 9B \\
       \hline
    \end{tabular}
    \label{tab:llms}
\end{table}

We evaluate the effectiveness of two components in CoM: methodology selection and methodology-guided reasoning. 

\noindent \textbf{LLMs} As listed in Table~\ref{tab:llms}, we report experiment results conducted on a relatively large and small LLMs as well as a recent open-source model reminiscent of OpenAI's o1, named Macro-o1, which is a fine-tuned Qwen2-7B-Instruct with a combination of the filtered Open-O1 CoT dataset \cite{Open-O1-CoT}, Macro-o1 CoT dataset, and Macro-o1 Instruction dataset. We use the LLM API provided by Siliconflow \cite{siliconflow} and Baidu Cloud \cite{cloudbaidu}, with settings: max\_tokens=1024, temperature=0.2, top\_k=40, top\_p=0.95, n=1.

\noindent \textbf{Dataset} We evaluate CoM using the same methodology definitions (Figure~\ref{fig:methodologies}) on the test splits of the datasets listed in Table~\ref{tab:datasets}.

\textit{AIME}: The 1983-2024 part of the American Invitational Mathematics Examination, includes complex algebraic equations, geometric puzzles, and advanced number theory problems to assess mathematical understanding and problem-solving skills.

\textit{GSM8K}: Linguistically diverse grade school math word problems requiring 2 to 8 steps of elementary calculations to solve.

\textit{MATH-500}: 500 problems from the MATH benchmark created by OpenAI.

\textit{HotpotQA}: The hard portion of a multi-hop, multi-step QA dataset. We simulate retrieval-augmented generation \cite{rag} by presenting only the question to LLMs. When LLMs generate code calling a search with keywords, we use fuzzy string matching to retrieve the top-k most similar supporting facts.

\textit{ARC}: AI2’s Reasoning Challenge dataset, which is a multiple-choice QA dataset with science exam questions from grades 3 to 9.

\begin{table}[t]
    \small
    \centering
    \caption{Datasets used in our experiments.}
    \begin{tabular}{l r} 
       \hline
       \textbf{Dataset} & \textbf{Size} \\ 
       \hline
       AIME \cite{aime} & 933 \\ 
       GSM8K \cite{gsm8k} & 1319 \\
       MATH-500 \cite{MATH-500} & 500 \\
       ARC \cite{arc} & 1172 \\
       HotpotQA \cite{glm2024chatglm} & 100 \\
       \hline
    \end{tabular}
    \label{tab:datasets}
\end{table}

\subsection{Baselines}

We evaluate CoM using zero-shot prompting, as few-shot approaches rely on task-specific examples, making them unsuitable for cross-domain comparisons. We compare CoM with three baselines that use recent prompting techniques and the same methodology definitions (Figure~\ref{fig:methodologies}), as well as Macro-o1 \cite{marco-o1}, a recent open-source model similar to OpenAI o1.

\textbf{CoT} \cite{wei2022chain} prompts the LLM to generate a chain of reasoning steps and shows the final result format.

\textbf{MCoT} provides the same methodology definitions as CoM, along with a CoT instruction to guide the LLM in using these methodologies in an appropriate order. MCoT evaluates whether methodologies can enhance reasoning in a single-turn, non-interactive setting, drawing on ideas from Least-to-Most \cite{zhou2023leasttomost} and Metacognitive-Prompting \cite{metacognitive-prompt}.

Both CoT and MCoT prompt the LLM once and do not allow code generation, as a second prompt is needed to synthesize code output.

\textbf{Workflow} is similar to CoM but uses a fixed methodology sequence per task, derived from the most frequent sequences chosen by CoM (Table~\ref{tab:sequences}). It guides the LLM through multiple reasoning turns, with sequences [Analysis, Coding, Variation, Conclusion] for AIME, GSM8K, and MATH, and [Analysis, Retrieval, Conclusion] for Hard Hotpot and ARC. Workflow incorporates ideas from Program-of-Thoughts \cite{PoT}, Cognitive Prompting \cite{metacognitive-prompt}, workflow/pipeline \cite{disentangle, CoI}, and RAG \cite{rag}.

\subsection{Performance Comparison}

Using the large LLM Qwen2-72B-Instruct, Figure~\ref{fig:qwen-large} shows that CoM outperforms baselines on AIME and Hard Hotpot, with accuracy and F1 improvements of 38.5\% and 28.7\%, respectively, over CoT. Results are similar for DeepSeek-V3 in Figure~\ref{fig:deepseek}. However, CoT slightly outperforms CoM on GSM8K, likely due to its simplicity and benchmark leakage \cite{xu2024benchmarking}. Workflow, which is task-specifically optimized, ranks second, while MCoT results suggest minimal benefits from single-prompt methodologies.

CoM’s performance is sensitive to the LLM’s meta-cognitive ability (i.e., its capacity to select methodologies dynamically). While Workflow benefits from task-specific prompting (its subtasks align with the most frequent successful sequences chosen by CoM, as shown in Table 4), CoM remains task-agnostic and thus more generalizable. Importantly, CoM outperforms Workflow on complex tasks (e.g., AIME with Qwen2-72B-Instruct and DeepSeek-V3) and on 4 out of 5 tasks with the smaller Qwen2.5-7B-Instruct (Figure \ref{fig:qwen-small}). This suggests that dynamic selection becomes increasingly advantageous as task complexity or model capability grows.

As shown in Figure~\ref{fig:qwen-large}, when compared with the task-specifically optimized Workflow, CoM's accuracy is 1.7\% higher on AIME, and 9.8\% higher on Hard Hotpot, demonstrating CoM's superior flexibility in methodology selection. This highlights the effectiveness of metacognitive abilities in LLMs for choosing appropriate methodology sequences and validates our step-by-step reasoning approach.

With the smaller LLM Qwen2.5-7B-Instruct (Figure~\ref{fig:qwen-small}), CoM remains the best performer on AIME, MATH, and ARC. Likely due to benchmark leakage \cite{xu2024benchmarking}, both CoM and Workflow show lower accuracy. On Hard Hotpot, CoM slightly underperforms Workflow, suggesting weaker metacognitive abilities in smaller models for methodology selection.

CoM's effectiveness depends on the LLM's metacognitive capabilities. As demonstrated in our Appendix Figures~\ref{fig:yi} to \ref{fig:glm}, models lacking these abilities struggle with methodology selection.However, Figure~\ref{fig:qwen-small} shows that Qwen2.5-7B-Instruct (7B parameters) consistently outperforms baselines across multiple tasks. This suggests that metacognitive abilities may emerge at smaller scales.

Finally, we compare CoM with Macro-o1 \cite{marco-o1} in Table~\ref{tab:macro-o1}. Results reveal that fine-tuning fails to improve Macro-o1 on AIME and Hard Hotpot, indicating insufficient generality in the fine-tuning data. For Macro-o1, we only evaluated single-round methods (CoT/MCoT) because it lacks multi-turn instruction-following capability.

\begin{table}[t]
    \centering
    \small
    \caption{Performance of Macro-o1 and Qwen2.5-7B-Instruct on AIME Tasks and Hard Hotpot}
    \begin{tabular}{l c | c c c c}
       \hline
        & \textbf{AIME} & \multicolumn{4}{c}{\textbf{Hard Hotpot}} \\ 
        & \textbf{Acc} & \textbf{EM} & \textbf{F1} & \textbf{Prec} & \textbf{Rec} \\ 
        \hline
        \multicolumn{2}{l}{\textbf{Macro-o1}} & & & &  \\
        CoT & 14.47 & 0.12 & 0.20 & 0.20 & 0.27 \\ 
        MCoT & 10.50 & 0.09 & 0.19 & 0.19 & 0.25 \\ 
        \hline
        \multicolumn{6}{l}{\textbf{Qwen2.5-7B-Instruct}} \\
        CoT & 20.15 & 0.17 & 0.25 & 0.25 & 0.24 \\ 
        MCoT & 17.58 & 0.15 & 0.22 & 0.22 & 0.22 \\ 
        CoM & \underline{25.4} & \underline{0.33} & \underline{0.42} & \underline{0.42} & \underline{0.51} \\ 
        % Workflow & 23.58 & 0.35 & 0.439 & 0.437 & 0.5751 \\ 
        \hline
    \end{tabular}
    \label{tab:macro-o1}
\end{table}

\subsection{Methodology Selection Patterns}

\begin{table}[t]
    \small
    \centering
    \caption{Top 52.2\% selected methodology sequences on AIME}
    \begin{tabular}{c p{0.37\textwidth}} 
       \hline
       \textbf{} & \textbf{Methodology Sequence} \\ 
       \hline
       22.0\% & Analysis Coding Validation Conclusion \\ 
       16.2\% & Analysis Coding Conclusion \\ 
       5.4\% & Analysis Coding Validation Reflection Flexibility Conclusion \\ 
       4.4\% & Analysis Coding Validation Reflection Conclusion \\ 
       4.3\% & Analysis Coding Validation Reflection Flexibility Validation Conclusion \\ 
       \hline
    \end{tabular}
    \label{tab:sequences}
\end{table}

We analyzed the reasoning history from the experiment to identify the most frequent methodology sequences selected by CoM. Table~\ref{tab:sequences} presents the top five patterns, which account for 52\% of CoM's responses using Qwen2-72B-Instruct.

In 22.0\% of cases, CoM followed a structured approach: analyzing, generating and executing code, validating results, and drawing conclusions. In 16.2\% of cases, the model skipped validation, suggesting high confidence in its code. The remaining patterns involved additional error correction steps, indicating potential validation issues. These findings suggest that LLMs exhibit metacognitive abilities by planning their reasoning steps during problem-solving.

\subsection{Ablation Study}

\begin{table}[t]
    \small
    \centering
    \caption{Ablation Study Results for COM Method}
    \begin{tabular}{lcccc}
        \hline
        \textbf{CoM} & \textbf{AIME (\%)} & \textbf{Hard Hotpot} \\
        \hline
        No Ablation & \underline{25.4} & \underline{0.4174} \\
        - Interpreter & 14.1 (-44.5\%) & 0.25 (-40.2\%) \\
        - Analysis & 18.7 (-26.6\%) & 0.38 (-8.4\%) \\
        - Coding & 23.3 (-8.3\%) & 0.38 (-8.8\%) \\
        - Retrieval & - & 0.22 (-46.8\%) \\
        - Validation & 23.9 (-7.2\%) & 0.4 (-3\%) \\
        - Reflection & 22.8 (-10.5\%) & 0.38 (-9\%) \\
        - Synthesis & 23 (-9.3\%) & 0.4 (-3\%) \\
        \hline
    \end{tabular}
    \label{tab:ablation}
\end{table}

We study the relative importance of each component of our CoM, including the Python interpreter, and each of the methodologies we used. Here, in contrast to excluding the \emph{code} methodology, which prevents CoM from generating code, removing the Python interpreter still allows the LLM to generate code, but the LLM then needs to guess the output of the code by itself without an interpreter. Our ablation study is conducted with Qwen2.5-7B-Instruct.

As listed in Table~\ref{tab:ablation}, the interpreter is very important for both tasks, which shows that the code output guessed by the LLM without using the code interpreter for both math calculation and knowledge retrieval is unreliable. Secondly, for hard math problems in AIME, systematic analysis of the data and constraints in the problem is vital for the correctness of the reasoning. For AIME, all methodologies we provided are useful, each contributing to a 7-10\% improvement in accuracy. In AIME, we disable retrieval for experimental simplicity. For Hard Hotpot, where reasoning relies on retrieved information, retrieval is clearly the most important methodology.

\subsection{Error Analysis}

Errors made in CoM are conventional LLM errors such as hallucination, misunderstanding, and instruction-following errors. We manually inspected the first 10 error cases in CoM on the GSM8K dataset. We found that in most of these cases, methodology selection is not perfect. Three error cases are due to hallucination, where the wrong answers are given directly without the necessary calculation process. Two cases are due to translation errors from natural language to math; for example, ``born early'' is translated to a reduction in age. Three cases are due to language understanding errors; for instance, ``restart downloading'' is understood as ``continue downloading'', and ``every second'' is understood as ``from the second''. In one error case, the initial calculation is correct, but then a validation step causes an error because the LLM believes ``servings'' must be an integer. In one error case, the LLM generates more than one code block, although the methodology definition contains an instruction to generate a single standalone code block.

\subsection{Efficiency}

\begin{table}[t]
    \small
    \centering
    \caption{Average Speed of Experiments in Seconds per Iteration (Multiplied by 50)}
    \begin{tabular}{l  c  c c}
        \hline
        & \textbf{AIME} & \textbf{GSM8K} & \textbf{Hard Hotpot} \\ 
        \hline
        \multicolumn{4}{l}{\textbf{Macro-o1}} \\
        CoT         & 96.0 & 33.5 & 19.0 \\ 
        MCoT        & 84.5 & 42.5 & 20.0 \\ 
        \hline
        \multicolumn{4}{l}{\textbf{Qwen2.5-7B-Instruct}}  \\
        CoM         & 91.0 & 34.0 & 50.5 \\ 
        Workflow    & 36.0 & 18.0 & 21.5 \\ 
        CoT         & 19.0 & 5.0 & 3.5 \\ 
        MCoT        & 19.0 & 8.0 & 3.5 \\ 
        \hline
    \end{tabular}
    \label{tab:speed}
\end{table}

We examine the inference efficiency in terms of total inference time for speed and the number of inferences for cost. Table~\ref{tab:speed}, shows that the speed of CoM is around 5 times that of CoT in AIME and 7 times in Hard Hotpot. However, CoM is comparable to the fine-tuned model in terms of total runtime. This is because total runtime is dominated by completion tokens, and CoM’s multi-turn calls (2–16 short prompts) vs. Macro-o1’s single long call. Latency is comparable because CoM’s prompts are concise.

Com strikes a good balance between performance and efficiency. Regarding performance, CoM shows significant gains over CoT (e.g., +25\% on AIME, +7\% on MATH, +14\% on ARC). While improvements over Workflow are narrower, Workflow is a highly optimized, task-specific baseline—making CoM’s competitive performance noteworthy. Regarding efficiency, although CoM uses ~10 prompts per question (Table \ref{tab:cost}), each prompt elicits short responses (Figure \ref{fig:select}), keeping latency comparable to fine-tuned models (Table \ref{tab:speed}) and far lower than search-based methods (e.g., ToT).

Table~\ref{tab:cost} compares the number of prompts made by CoM with those made by Workflow. The results show that although we set the maximum iteration $K=8$, CoM stops at a smaller number of steps than the maximum iterations 2*K on average, generates more reasoning steps for the harder AIME problems, and a smaller number of steps for the easier GSM8K problems.

\subsection{Summary of Experiments}

The experiments on complex mathematical problems (AIME and GSM8K) and multi-hop question answering (HotpotQA) evaluate the effectiveness of CoM in methodology selection and guided reasoning using a 72B, a 7B LLM, and a fine-tuned LLM for structured reasoning.

Results show that our CoM is effective in improving the performance of two challenging tasks over baselines that embody recent prompt engineering approaches. This result supports our hypothesis that we can use a training-free solution that integrates human methodological insights to enhance the performance of LLMs in complex reasoning.

Methodology selection patterns reveal that CoM effectively generates reasonable methodology sequences, which guide its reasoning in the right direction. Error analysis identifies that common LLM issues contribute to the majority of errors made by CoM. Finally, the ablation study confirms that the methodologies we employed are critical for solving complex reasoning tasks.

% \begin{table}[t]
%     \centering
%     \caption{Average Speed of Experiments in Iterations per Second}
%     \begin{tabular}{l  c  c c}
%         \hline
%         & \textbf{AIME} & \textbf{GSM8K} & \textbf{Hard Hotpot} \\ 
%         \hline
%         \multicolumn{4}{l}{\textbf{Macro-o1}} \\
%         CoT         & 0.52 & 1.49 & 2.63 \\ 
%         MCoT        & 0.59 & 1.18 & 2.50 \\ 
%         \hline
%         \multicolumn{4}{l}{\textbf{Qwen2.5-7B-Instruct}}  \\
%         CoM         & 0.55 & 1.47 & 0.99 \\ 
%         Workflow    & 1.38 & 2.80 & 2.33 \\ 
%         CoT         & 2.60 & 9.74 & 14.29 \\ 
%         MCoT        & 2.60 & 6.07 & 14.29 \\ 
%         \hline
%     \end{tabular}
%     \label{tab:average_speed}
% \end{table}
        
\begin{table}[t]
    \small
    \centering
    \caption{Average Number of Prompts (not tokens) per Question}
    \begin{tabular}{l c c c}
        \hline
        & \textbf{AIME} & \textbf{GSM8K} & \textbf{Hard Hotpot} \\ 
        \hline
        CoM         & ${2\times}$5.76 & ${2\times}$3.99 & ${2\times}$5.98 \\ 
        Workflow    & 4 & 4 & 3 \\ 
        \hline
    \end{tabular}
    \label{tab:cost}
\end{table}

\section{Conclusion and Future Work}

This paper enhances LLMs' reasoning capabilities for complex tasks by simulating metacognitive processes and leveraging user-defined methodologies, enabling effective navigation of complex reasoning tasks without extensive retraining. Takeaways include: (1) LLMs exhibit latent metacognitive abilities that can be activated through structured, justification-driven prompting--eliminating the need for fine-tuning; and (2) generating explicit methodology justifications improves traceability and task comprehension, boosting zero-shot accuracy and cross-domain adaptation, which is often constrained by limited in-context examples.

he primary objective of this paper was to investigate the feasibility of using LLM self-generated thought guidelines (e.g., methodologies) to steer its reasoning process. Future work includes fine-tuning a small adaptor to improve metacognitive capabilities and shifting the design effort from prompt frameworks to the engineering of methodologies.

% A promising direction for future work is the automated search for optimal methodologies and reasoning chains, inspired by approaches like \cite{llm-optim}, to identify and apply methodologies that enhance reasoning performance through better prompt construction.

\section*{Limitations}

The approach proposed in this paper assumes that the LLM possesses metacognitive abilities. We found that other LLMs, despite demonstrating competitive performance in various benchmarks, fail in methodology selection, even with extensive prompt tuning efforts. For instance, one of these LLMs consistently selects the first methodology it initially chose. Additional experimental results illustrating these failures are provided in Appendix~\ref{appendix:fail}.

Currently, our method requires two distinct prompts at each step: one for methodology selection and another for methodology-based reasoning. We attempted to consolidate these two prompts into a single one; however, we observed that even the most advanced LLMs we tested, including DeepSeek-V3, struggled to follow instructions with the combined, more complex prompt. We anticipate that future advancements in LLMs' instruction-following capabilities will enable the use of a single prompt, thereby improving the efficiency of our method.

The methodologies included in our framework are not exhaustive, leaving room for future research to expand and refine the list. Incorporating a wider range of strategies could enhance the adaptability and robustness of the CoM framework, opening new avenues for exploration and improvement.

\section*{Ethical Statement}

This work fully complies with the ACL Ethics Policy. We declare that there are no ethical issues in this paper, to the best of our knowledge.

\bibliography{ref}

%\newpage
\appendix

\section{Appendix}

\subsection{Prompts} \label{appendix:prompts}

\begin{figure*}
  \centering
  \includegraphics[trim=45 180 448 80,clip,width=.9\linewidth]{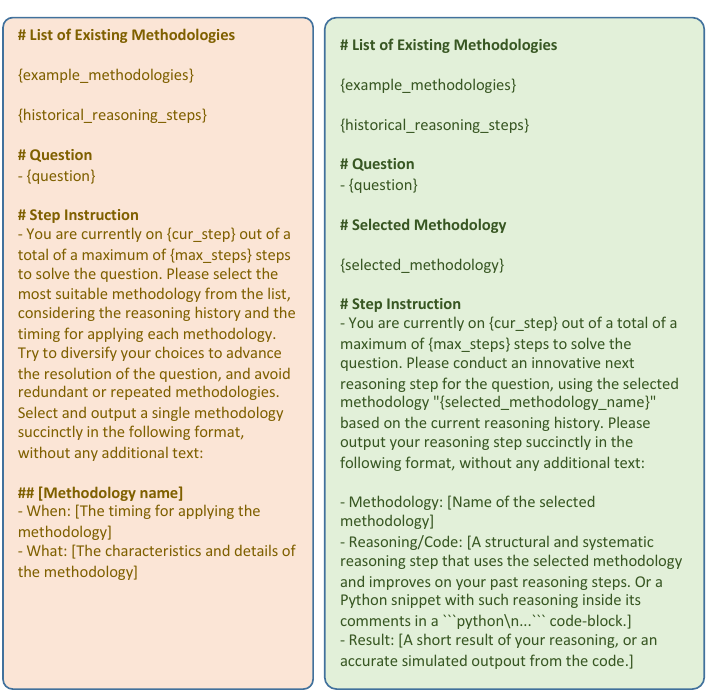}
  \caption{Our methodology-selection prompt (left) and methodology-based reasoning prompt (right).}
  \label{fig:select}
\end{figure*}

\begin{figure*}
  \centering
  \includegraphics[trim=45 90 460 80,clip,width=.9\textwidth]{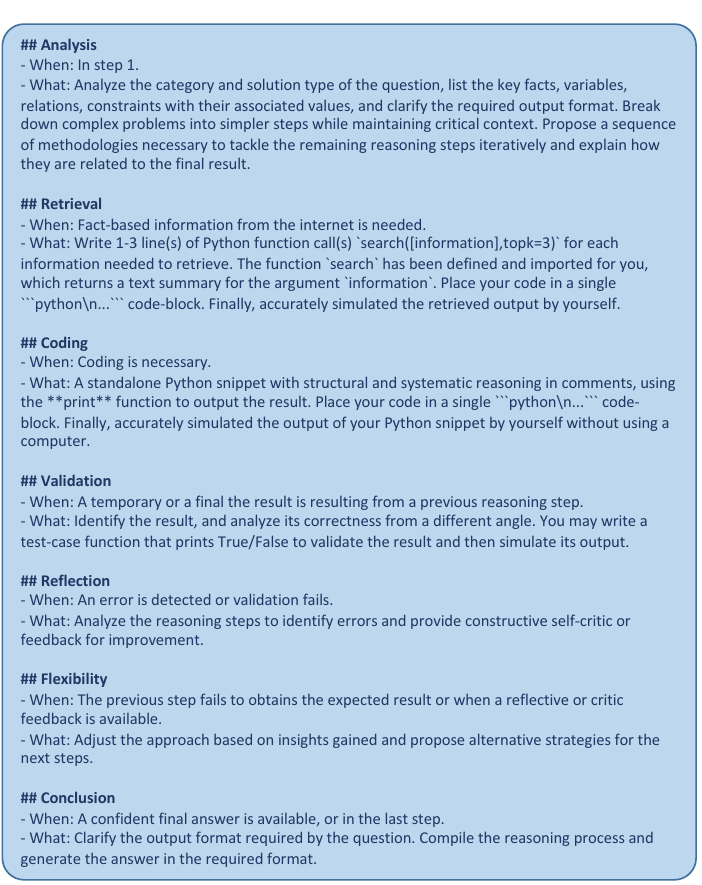}
  \caption{Our list of methodologies.}
  \label{fig:methodologies}
\end{figure*}

Our list of methodologies is displayed in Figure~\ref{fig:methodologies}, and our methodology-selection and methodology-based reasoning prompts are listed in Figure~\ref{fig:select}.

\subsection{Additional Experiment Results} \label{appendix:fail}

\begin{figure*}[ht]
\begin{subfigure}{.32\textwidth}
  \centering
  \includegraphics[width=\linewidth]{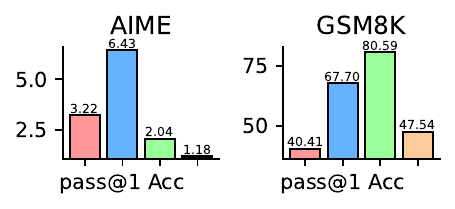}  
\end{subfigure}
\begin{subfigure}{.68\textwidth}
  \centering
  \includegraphics[width=\linewidth]{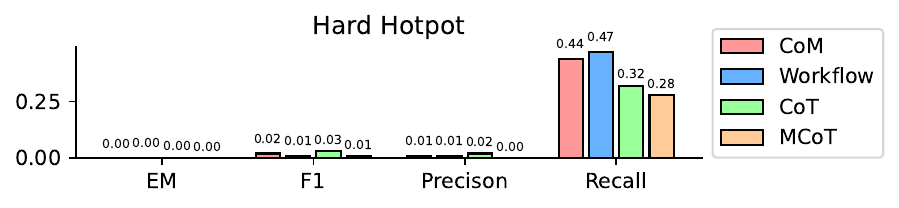}  
\end{subfigure}
\vspace{-20pt}
\caption{Results of Yi-1.5-9B-Chat \cite{yi} on AIME, GSM8K, and Hard HotpotQA.}
\label{fig:yi}
\end{figure*}

\begin{figure*}[ht]
\begin{subfigure}{.32\textwidth}
  \centering
  \includegraphics[width=\linewidth]{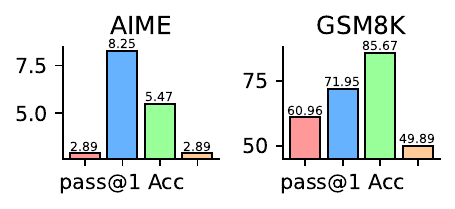}  
\end{subfigure}
\begin{subfigure}{.68\textwidth}
  \centering
  \includegraphics[width=\linewidth]{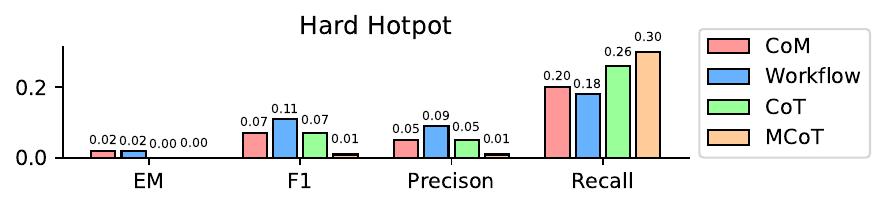}
\end{subfigure}
\vspace{-20pt}
\caption{Results of InternLM2.5-7B-chat \cite{cai2024internlm2} on AIME, GSM8K, and Hard HotpotQA.}
\label{fig:intern}
\end{figure*}

\begin{figure*}[ht]
\begin{subfigure}{.32\textwidth}
  \centering
  \includegraphics[width=\linewidth]{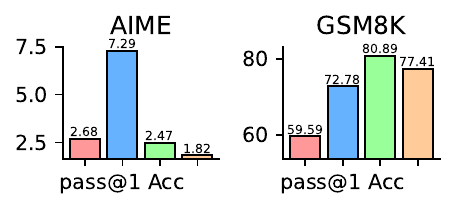}  
\end{subfigure}
\begin{subfigure}{.68\textwidth}
  \centering
  \includegraphics[width=\linewidth]{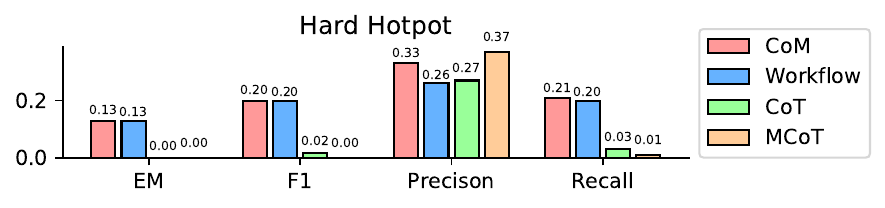}  
\end{subfigure}
\vspace{-20pt}
\caption{Results of GLM-4-9b-chat \cite{glm2024chatglm} on AIME, GSM8K, and Hard HotpotQA.}
\label{fig:glm}
\end{figure*}

The approach proposed in this paper assumes that the LLM possesses metacognitive abilities. This section presents additional experimental results in Figures~\ref{fig:yi}, \ref{fig:intern}, and \ref{fig:glm}, which reveal that some LLMs, despite demonstrating competitive performance across various benchmarks, struggle with methodology selection even after extensive prompt tuning efforts. For example, one of these LLMs consistently defaults to selecting the first methodology it initially identifies, highlighting a limitation in its decision-making process.

We ran experiments with Self-Consistency \cite{wang2023selfconsistency} (CoT-SC) as shown in Table~\ref{fig:cot-sc}. Note that CoT-SC improves over CoT but SC is orthogonal to CoM.

\begin{table}[h]
    \small
    \centering
    \caption{Performance Results of Different Methods}
    \begin{tabular}{l c c c}
        \hline
        \textbf{Method} & \textbf{AIME} & \textbf{GSM8K} & \textbf{MATH} \\ 
        \hline
        CoT            & 20.15 & 91.51 & 55.06 \\ 
        CoT-SC         & 27.12 & 92.19 & 68.00 \\ 
        CoM            & 25.40 & 84.53 & 58.80 \\ 
        \hline
    \end{tabular}
    \label{tab:results}
\end{table}

\end{document}